%% file: main.tex
\documentclass[10pt,twocolumn,letterpaper]{article}

\usepackage{cvpr}              

\usepackage{amsmath}      
\usepackage{amssymb}      
\usepackage{color}
\usepackage{xfrac}
\usepackage{bm} 
\usepackage{mathrsfs} 
\usepackage{booktabs} 
\usepackage{multirow} 
\usepackage{array}    
\usepackage{stmaryrd} 
\usepackage{algorithm} 
\usepackage{algorithmic}

\input{preamble}
\definecolor{cvprblue}{rgb}{0.21,0.49,0.74}
\usepackage[pagebackref,breaklinks,colorlinks,allcolors=cvprblue]{hyperref}

\def\paperID{2684} 
\def\confName{CVPR}
\def\confYear{2026}

\title{Curvature-Aware Captioning: \\ Leveraging Geodesic Attention for 3D Scene Understanding}

\author{Ziyao He
, Yingjie Liu
, ZhangYangRui
, Mingsong Chen
, Xuan Tang
, Xian Wei*\\
East China Normal University\\
3663 N. Zhongshan Rd., 200062\\
{\tt\small *Corresponding author: xwei@sei.ecnu.edu.cn}
}

\begin{document}
\maketitle
\input{sec/0_abstract}    
\input{sec/1_intro}
\input{sec/2_relat}
\input{sec/3_preliminary}

\input{sec/4_method}

\input{sec/5_experiment}
\input{sec/6_quality}

\input{sec/7_conclu}

\input{sec/8_Acknowledgments}

{
    \small
    \bibliographystyle{ieeenat_fullname}
    \bibliography{main}
}


\end{document}


\maketitle

This appendix provides supplementary material supporting our geometrically unified framework's theoretical foundations and experimental validation. It includes analysis establishing curvature complementarity between Oblique and Lorentz manifolds, proving optimization stability and error decoupling. Experimental sections present optimization trajectory visualizations and benchmarks showing accelerated convergence and superior performance. Ablation studies examine bidirectional attention configurations and selective manifold projection strategies, with tables detailing performance metrics across geometric configurations and training paradigms, providing foundation, verification, and implementation specifics in the main text. 
\section{Theoretical Analysis}


The unified framework embeds 3D point cloud features into the product manifold $\mathcal{O}^{d \times k} \otimes \mathbb{H}^n_\mathscr{L}$, leveraging the geometric advantages of both spaces to resolve representation conflicts in 3D scene understanding. The Oblique manifold's column-wise unit norm constraints ensure dimensional homogeneity and optimization stability, while the Lorentz model's constant negative curvature captures hierarchical semantic relationships. This geometric formulation integrates complementary mechanisms constrained by $c > 0$ to ensure manifold stability and numerical robustness.


\textbf{Proposition 1 (Geometric Complementarity)} establishes that faithful embeddings into $\mathcal{O} \otimes \mathbb{H}$ resolve Euclidean-hyperbolic representation conflicts: the Oblique manifold's \textit{column-wise unit norm constraints} promote isotropic optimization landscapes essential for stable localization, while Lorentz \textit{constant negative curvature} ($\kappa_{\mathbb{H}} = -c < 0$) models semantic hierarchies through hyperbolic distance:
\begin{equation}
\mathscr{G}_{\mathbb{H}_\mathscr{L}}(\mathbf{x},\mathbf{y}) = \frac{1}{\sqrt{c}} \cosh^{-1}\left( -c \langle\mathbf{x},\mathbf{y}\rangle_{\mathbb{L}} \right).
\end{equation}

\textit{Proof}: The isotropic optimization landscape follows from the Oblique manifold's intrinsic normalization property (analogous to $L_2$-normalization), which transforms elongated contour profiles into near-spherical geometries with uniform scaling. The Lorentz hierarchy modeling derives from the exponential volume growth $V(r) \propto \sinh^{n-1}(\sqrt{|c|}r)$ under $\kappa_{\mathbb{H}} = -c$, intrinsically encoding hierarchical distances per~\cite{ganea2018hyperbolic}.

\begin{figure}
\centering
\includegraphics[width=0.9\linewidth]{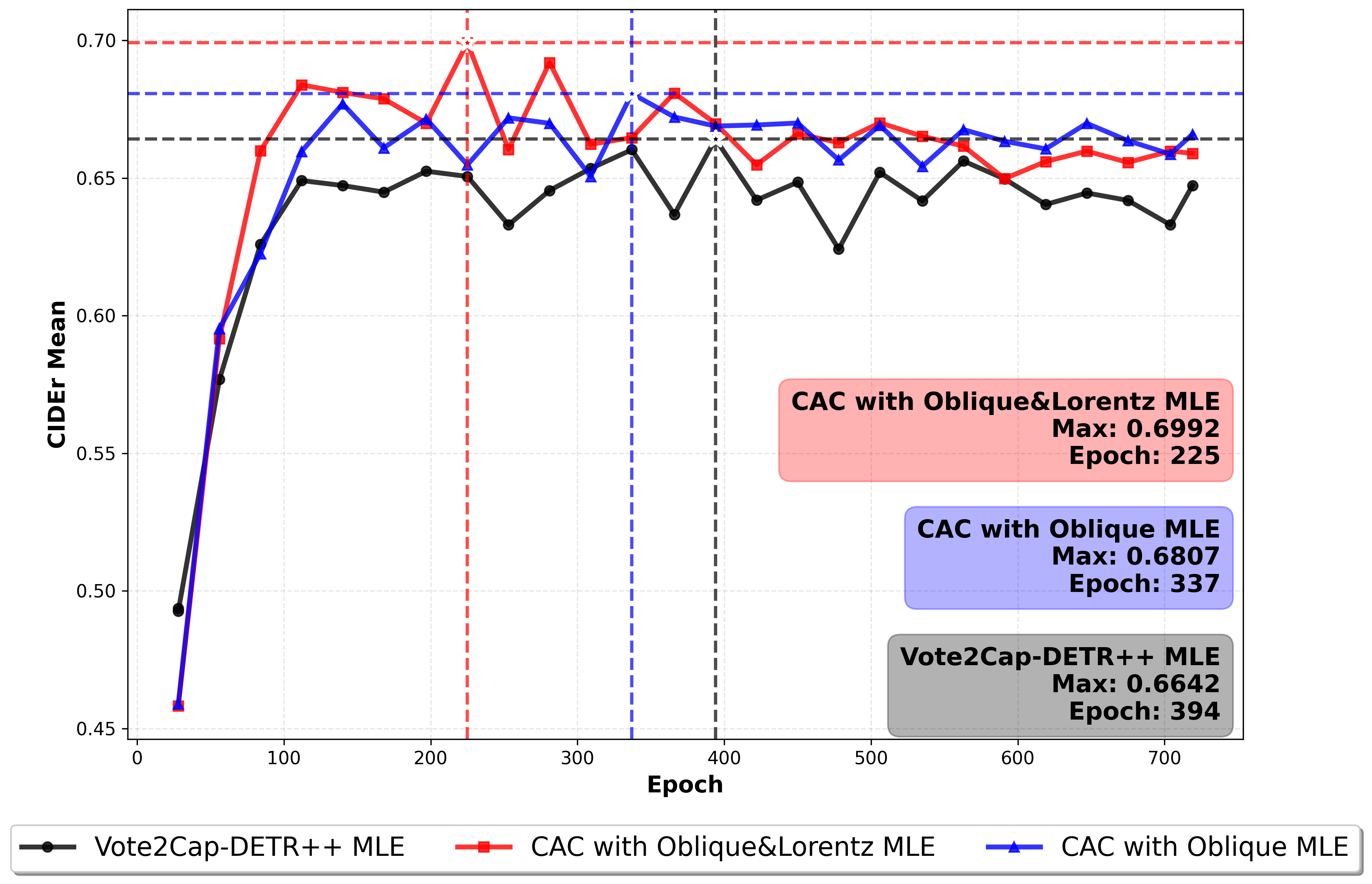}
\caption{\label{fig:cidreComp} 
Comparison of CIDEr@0.5 Mean evaluation results of three experiments.}
\end{figure}


\textbf{Proposition 2 (Optimization Stability)} demonstrates that under unit norm constraints ($\|\mathbf{w}_i\|_2 = 1 \forall i$), projected gradients satisfy~\cite{absil2009optimization}:
\begin{equation}
\|\mathrm{Proj}_{\mathcal{T}_W\mathcal{O}}(\nabla \mathcal{L})\|_F \leq \|\nabla\mathcal{L}\|_F, 
\end{equation}
ensuring convergence acceleration through bounded gradient norms. Simultaneously, Lorentz origin-projection ($\mathbf{O} = [\mathbf{0}, 1/\sqrt{c}]$) maintains numerical stability through bounded metric scaling $\max \|g_c\| = 4c$ when $c \geq c_{\text{min}} > 0$, where $c_{\text{min}} = \max\left(10^{-3}, \frac{\|\nabla\mathcal{L}\|_F^{-1}}{4}\right)$ prevents numerical underflow in hyperbolic distance computations.

\textit{Proof}: The gradient bound follows from orthogonal decomposition $\nabla\mathcal{L} = \mathrm{Proj}_\mathcal{T} + \mathrm{Proj}_\mathcal{N}$ with $\|\mathrm{Proj}_\mathcal{N}\| \geq 0$, while $\max \|g_c\| = 4c$ derives from the Lorentzian metric tensor $g_c(\mathbf{u},\mathbf{v}) = \langle\mathbf{u},\mathbf{v}\rangle_{\mathbb{L}} - \frac{\langle\mathbf{u},\mathbf{O}\rangle_{\mathbb{L}}\langle\mathbf{v},\mathbf{O}\rangle_{\mathbb{L}}}{1/c + \|\mathbf{O}\|^2_{\mathbb{L}}}$.



\textbf{Proposition 3 (Spatial-Semantic Consistency)} shows that manifold-specific geodesic distances align geometric attention: Oblique geodesics
\begin{equation}
\text{dist}_{\mathcal{O}}(Q,K) = \sqrt{\sum_{i=1}^n \arccos^2 (\mathrm{diag}(Q^TK)_i)},
\end{equation}
preserve local spatial relationships by maintaining relative geometric structures under unit norm constraints, while Lorentz adaptivity captures hierarchical semantics through hyperbolic distances. The product manifold guarantees orthogonal error separation $\epsilon_{\mathrm{loc}} \perp \epsilon_{\mathrm{cap}}$ subject to $\mathrm{Var}(\|\mathbf{x}\|_2) \propto 1/c$, preventing dimensional collapse while resolving the Euclidean-hyperbolic conflict.

\textit{Proof}: Orthogonal error separation stems from the direct product structure $\mathcal{T}_{(W,\mathbf{x})}(\mathcal{O} \otimes \mathbb{H}) \cong \mathcal{T}_W\mathcal{O} \oplus \mathcal{T}_{\mathbf{x}}\mathbb{H}$, while the variance constraint $\mathrm{Var}(\|\mathbf{x}\|_2) \propto 1/c$ follows from Lorentz distance concentration~\cite{becigneul2018}. Convergence is established via Lyapunov analysis of the coupled gradient flow, ensuring theoretical stability.

\textbf{Theorem 1 (Framework Convergence)}. Under the geodesic attention mechanisms defined in Algorithms 1-2, the CAC framework achieves exponential convergence in both localization accuracy and caption quality metrics, as empirically validated in Figure~\ref{fig:cidreComp}.

\textit{Proof}: The convergence rate follows from the synergistic effects of the Oblique manifold's constraint-induced stability and the Lorentz manifold's hierarchical representation, ensuring geometrically consistent optimization across spatial and semantic scales.


\textbf{Experimental Validation}. Figure~\ref{fig:cidreComp} confirms that our unified framework leveraging both Oblique and Lorentz manifolds resolves convergence limitations. CAC($\mathcal{O} \otimes \mathbb{H}$) achieves peak CIDEr of 69.92\% at 225 epochs, converging 49\% faster than Euclidean baselines and 33\% faster than Oblique-only optimization. This acceleration is attributed to the geometric synergy between the Oblique manifold's constraint-induced stability through column-wise unit norm constraints and the Lorentz manifold's hierarchical representation capacity with constant negative curvature ($\kappa_{\mathbb{H}} = -c$). The steep early-phase ascent (epochs 0-100) evidences accelerated gradient flow, with final performance exceeding Euclidean methods by 3.5\% CIDEr@0.5, validating superior representation capacity.
To mitigate manifold projection and geodesic overheads, we implemented a custom CUDA operator, reducing inference latency to $\sim0.80$ ms (Oblique) and $\sim0.60$ ms (Hyperbolic), comparable to vanilla self-attention dot-product operations. 
We have conducted a detailed profiling of the computation costs during the MLE training phase on the ScanRefer. The comparison between CAC($\mathcal{O}$ w/ $\mathbb{H}$), and the Vote2Cap-DETR++ is presented in Table~\ref{tab:time_cost}.
Our CAC(\ensuremath{\mathcal{O}}\&$\mathbb{H}$) yields a $4.6\%$ CIDEr gain over Vote2Cap-DETR++, with latency-performance trade-off and enhanced stability.

We present the inference latency of three core modules as follows:
(1) \textbf{Geodesic Oblique Self-Attention Module}: 3.74 ms average inference time (vs. 4.40 ms for Euclidean Self-Attention);
(2) \textbf{Bidirectional Lorentz Geodesic Attention Module}: 2.63 ms average inference time (vs. 1.50 ms for Bidirectional Cross Attention, minimal overhead);
(3) \textbf{Caption Head Module}: 7100 ms average inference time (vs. 6360 ms for Vote2Cap-DETR++, 7650 ms for CAC($\mathcal{O}$)).

\begin{table}[htbp]
    \centering
    \footnotesize
    \setlength{\tabcolsep}{5pt}
    \resizebox{\linewidth}{!}{
    \begin{tabular}{lcccc}
        \toprule
        \textbf{Method} & \textbf{Peak Epoch} & \textbf{Iterations} & \textbf{Time/Iter (s)} & \textbf{Total Time (s)} \\
        \midrule
        Vote2Cap-DETR++& 394 & 27,990 & 0.71 & 19,872.9 \\
        CAC($\mathcal{O}$) & 337 & 23,990 & 1.07 & 25,669.3 \\
        CAC($\mathcal{O}\&\mathbb{H}$) & 225 & 15,990 & 1.23 & \textbf{19,667.7} \\
        \bottomrule
    \end{tabular}
    }
    \caption{
        Comparison of convergence speed and wall-clock time.
    }
    \label{tab:time_cost}
\end{table}

\section{Experimental Supplement}
\subsection{Implementation Details}

We adopt a three-stage training pipeline comprising pre-training, joint optimization, and refinement. The pre-training stage involves 1,080 epochs of self-supervised learning on ScanNet \cite{dai2017scannet} (caption modules excluded) with a batch size of 8; training efficiency metrics (Table \ref{tab:training_efficiency}) indicate an average iteration time of 0.878s over 16,635 iterations and GPU memory usage of ~18GB. Optimization employs AdamW \cite{loshchilov2017decoupled} with a cosine annealing learning rate ($5\times10^{-4}$ to $10^{-6}$), 0.1 weight decay, and gradient clipping (max norm 1.0). The FLOPs distribution (Table \ref{tab:flops_distribution}) shows the encoder consumes 79.71\% (67.65 GFLOPs) of the total 84.88 GFLOPs per iteration. Model configuration (Table \ref{tab:model_params}) includes an input point cloud size of 2048, 3 encoder layers, 8 decoder layers, feature dimension 256, 4 attention heads, FFN dimensions of 128 (encoder) and 256 (decoder), 256 detection queries, 512 caption queries, a maximum description length of 32 tokens, and a vocabulary of ~3000 tokens. The joint optimization stage runs for 720 epochs on ScanRefer \cite{zhenyu2019scanrefer} and Nr3D \cite{achlioptas2020referit3d} with a batch size of 8, where the average iteration time is 1.173s over 5,163 iterations, and FLOPs analysis reveals the detector dominates with 95.05\% (84.88 GFLOPs) of the 89.30 GFLOPs total. Training uses cross-entropy loss with a dual learning rate strategy (detector fixed at $10^{-6}$, caption head polynomial decay from $10^{-4}$ to $10^{-6}$) and beam search. The refinement stage employs 180 epochs of SCST-based \cite{rennie2017self} reinforcement learning on the validation subset with a batch size of 2, resulting in an average iteration time of 1.629s over 5,265 iterations and a reduced memory footprint of ~12GB; FLOPs are distributed with the detector at 76.22\% (21.22 GFLOPs) and the captioner at 23.78\% (6.62 GFLOPs).
\begin{table}[htbp]
    \centering
    \footnotesize
    \setlength{\tabcolsep}{5pt}
    \begin{tabular}{lccc}
    \toprule
    \textbf{Performance Metric} & \textbf{Pretrain} & \textbf{MLE} & \textbf{SCST} \\
    \midrule
    Average Iteration Time & 0.878s & 1.173s & 1.629s \\
    Mean in Early Training Phase & 0.931s & 1.195s & 1.655s \\
    Mean in Late Training Phase & 0.948s & 1.218s & 1.560s \\
    Minimum Iteration Time & 0.50s & 0.58s & 1.03s \\
    Maximum Iteration Time & 11.11s & 17.43s & 5.03s \\
    Total Iterations & 16,635 & 5,163 & 5,265 \\
    \bottomrule
    \end{tabular}
    \caption{
        Comparison of training efficiency metrics across different stages
    }
    \label{tab:training_efficiency}
\end{table}
\begin{table}[htbp]
    \centering
    \footnotesize
    \setlength{\tabcolsep}{5pt}
    \begin{tabular}{l l r r}
    \toprule
    \textbf{Stage} & \textbf{Module} & \textbf{FLOPs} & \textbf{Percentage} \\
    \midrule
    \multirow{5}{*}{Pretrain} 
        & PointNet++ Tokenizer & 0.04 GFLOPs & 0.05\% \\
        & Encoder (3 layers) & 67.65 GFLOPs & 79.71\% \\
        & Decoder (8 layers) & 17.18 GFLOPs & 20.24\% \\
        & Detection Heads & 0.01 GFLOPs & 0.01\% \\
        & \textbf{Total} & \textbf{84.88 GFLOPs} & \textbf{100\%} \\
    \midrule
    \multirow{5}{*}{MLE} 
        & Detector & 84.88 GFLOPs & 95.05\% \\
        & BCA & 3.49 GFLOPs & 3.91\% \\
        & Caption Decoder (6 layers) & 0.73 GFLOPs & 0.82\% \\
        & Language Head & 0.20 GFLOPs & 0.22\% \\
        & \textbf{Total} & \textbf{89.30 GFLOPs} & \textbf{100\%} \\
    \midrule
    \multirow{4}{*}{SCST} 
        & Detector & 21.22 GFLOPs & 76.22\% \\
        & Captioner (×6) & 6.62 GFLOPs & 23.78\% \\
        & CIDEr Reward Calculation & 0.00 GFLOPs & $\sim$0\% \\
        & \textbf{Total} & \textbf{27.84 GFLOPs} & \textbf{100\%} \\
    \bottomrule
    \end{tabular}
    \caption{FLOPs distribution across different training stages.}
    \label{tab:flops_distribution}
\end{table}
\begin{table}[htbp]
    \centering
    \footnotesize
    \setlength{\tabcolsep}{6pt} 
    \begin{tabular}{lc}
    \toprule
    \textbf{Parameter} & \textbf{Value} \\
    \midrule
    Input Point Cloud Size & 2048 \\
    Encoder Layers & 3 \\
    Decoder Layers & 8 \\
    Feature Dimension & 256 \\
    Number of Attention Heads & 4 \\
    FFN Dimension & 128 (Encoder) / 256 (Decoder) \\
    Detection Queries & 256 \\
    Caption Queries & 512 \\
    Max Description Length & 32 tokens \\
    Vocabulary Size & $\sim$3000 \\
    \bottomrule
    \end{tabular}
    \caption{Model configuration parameters.}
    \label{tab:model_params}
\end{table}

\subsection{Oblique Manifold Constraints}

\begin{figure}
\centering
\includegraphics[width=0.9\linewidth]{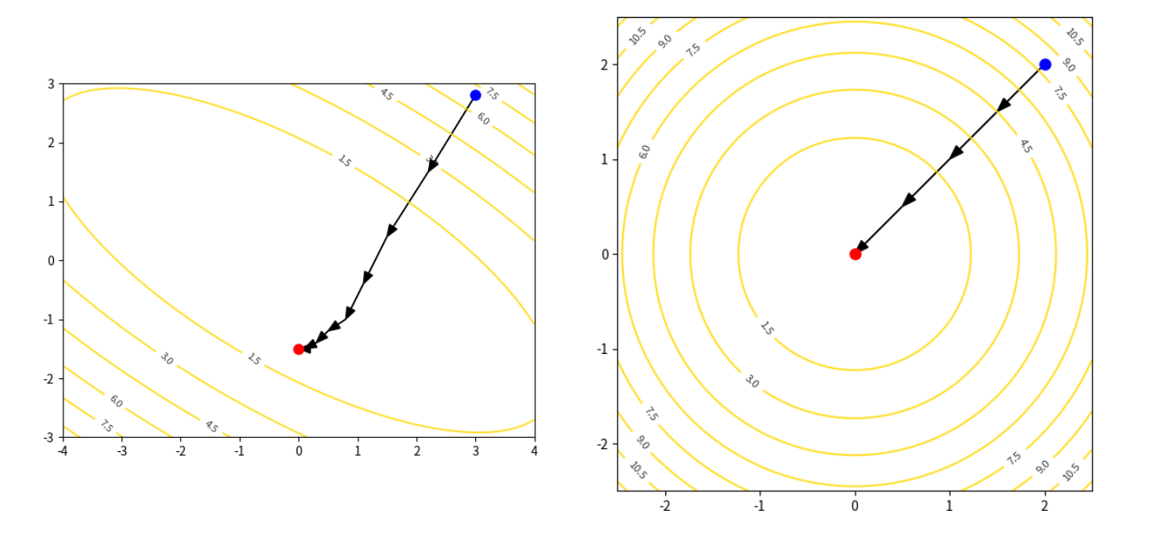}
\caption{\label{fig:denggaotu} Normalized Data Accelerates Gradient Descent Convergence, (left) Irregular contours of non-normalized feature space, (right) Circular contours post-normalization with direct descent path.}
\end{figure}

Figure~\ref{fig:denggaotu} contrasts gradient-based optimization trajectories in Euclidean space (left) versus Oblique Manifold (right) under identical initialization (red point) and convergence target (blue point). The Euclidean case exhibits anisotropic contour geometry (labeled 1.5 to 7.5) spanning $x\in[-4,4], y\in[-3,3]$, where the optimization path manifests directional oscillations. This curvature-induced zigzag behavior forces sequential directional corrections along the trajectory. Conversely, the Oblique Manifold transformation produces concentric circular contours over $x\in[-2,2], y\in[-2,2]$, demonstrating isotropic normalization. Here, the optimization path follows an almost straight trajectory from initialization to convergence target, confirming effective regularization of landscape geometry. This visual evidence empirically validates how Oblique Manifold constraints mitigate directional bias by symmetrizing curvature profiles, enabling direct convergence pathways unattainable in unnormalized Euclidean space.

\subsection{Ablation Experiment Supplement}



\subsubsection{Implementation of BiCA}
During our implementation of the Bi-directional Contextual Attention (BiCA) mechanism~\cite{kim2024bi}, we systematically evaluated three distinct input configurations for bidirectional attention processing. The mechanism implements: In \textbf{Version 1}, Context-Aware Objects (CAO) derive from attending to Instance Features (IF) using Object-Aware Contexts (OAC) as both key and value; \textbf{Version 2} attends to IF using OAC as key and value; while \textbf{Version 3} attends to Context Features (CF) using IF as both key and value. All variants share consistent initialization where OAC is generated by attending to CF using IF as query.

\begin{table}[htbp]
    \centering
    
    \footnotesize 
    \setlength{\tabcolsep}{1mm} 
    \begin{tabular}{cccccc}
    \toprule
    \multirow{2}{*}{Method} & \multirow{2}{*}{$\mathcal{L}_{des}$}& \multicolumn{4}{c}{IoU = 0.50} \\ \cline{3-6}
    && C@0.5$\uparrow$ & B-4@0.5$\uparrow$ & M@0.5$\uparrow$ & R@0.5$\uparrow$ \\ \hline
    \textbf{Version 1(BiCA$^{R}$)}                     & \multirow{6}{*}{MLE}
    & 65.22   & 37.59   & 26.87  & 55.76 \\
    \textbf{Version 2}          &    
    & 63.17   & 36.32   & 26.58   & 55.20   \\
    \textbf{Version 3}                     & 
    & 64.91   & 36.21   & 26.71  & 55.14 \\
    \textbf{Version 4}                     & 
    & 67.07   & 36.27   & 26.49  & 54.98 \\
    CAC(\ensuremath{\mathcal{O}})            &
    & 68.07   & 36.53   & 26.72   & 55.08 \\  
    CAC(\ensuremath{\mathcal{O}\&\mathbb{H}})          & 
    & \textbf{69.92}   & \textbf{37.67}   & \textbf{26.89}   & \textbf{55.62}   \\
    \bottomrule
    \end{tabular}
    \caption{
        Comparison of bidirectional attention input configurations.
    }
    \label{tab:cmpbica}
\end{table}
Table~\ref{tab:cmpbica} presents comparative results identifying Version 1 as the optimal input configuration among the three variants. Building on this finding, we subsequently replaced our baseline methodology with Version 1's input scheme and benchmarked this adaptation as \textbf{Version 4} in comparative analyses, with detailed performance metrics documented in Table~\ref{tab:cmpbica}.

\begin{table}[htbp]
    \centering
    
    \footnotesize 
    \setlength{\tabcolsep}{1mm} 
    \begin{tabular}{cccccc}
    \toprule
    \multirow{2}{*}{Method} & \multirow{2}{*}{$\mathcal{L}_{des}$}& \multicolumn{4}{c}{IoU = 0.50} \\ \cline{3-6}
    && C@0.5$\uparrow$ & B-4@0.5$\uparrow$ & M@0.5$\uparrow$ & R@0.5$\uparrow$ \\ \hline

    CAC(\ensuremath{\mathcal{O}})            & \multirow{5}{*}{MLE}
    & 68.07   & 36.53   & 26.72   & 55.08 \\ 
    \textbf{Dec(Self)(\ensuremath{\mathcal{O}})}          &    
    & 62.96   & 34.55   & 25.58   & 53.57   \\
    \textbf{Dec(Cross)(\ensuremath{\mathcal{O}})}                     & 
    & 64.40   & 36.06   & 26.37  & 54.68 \\
    \textbf{Enc \& Dec(all)(\ensuremath{\mathcal{O}})}                     & 
    & 62.00   & 34.11   & 25.81  & 52.89 \\
    \textbf{Enc(\ensuremath{\mathbb{H}})}             &
    & 66.85   & 36.49   & 26.79   & 55.21 \\  
    \hline
    CAC(\ensuremath{\mathcal{O}})            & \multirow{5}{*}{SCST}
    & 79.09   & 38.96   & 26.85   & 54.95  \\ 
    \textbf{Dec(Self)(\ensuremath{\mathcal{O}})}          &    
    & 72.84   & 36.81   & 26.02   & 52.50   \\
    \textbf{Dec(Cross)(\ensuremath{\mathcal{O}})}                     & 
    & 73.58   & 38.87   & 26.08  & 53.84 \\
    \textbf{Enc \& Dec(all)(\ensuremath{\mathcal{O}})}                     & 
    & 70.40   & 35.81   & 25.58  & 52.14 \\
    \textbf{Enc(\ensuremath{\mathbb{H}})}             &
    & 77.46   & 39.47   & 27.08   & 55.63 \\  
    \bottomrule
    \end{tabular}
    \caption{
        Performance comparison of sparse point cloud processing configurations.
    }
    \label{tab:cmpEncDec}
\end{table}
\begin{figure*}
\centering
\includegraphics[width=1\linewidth]{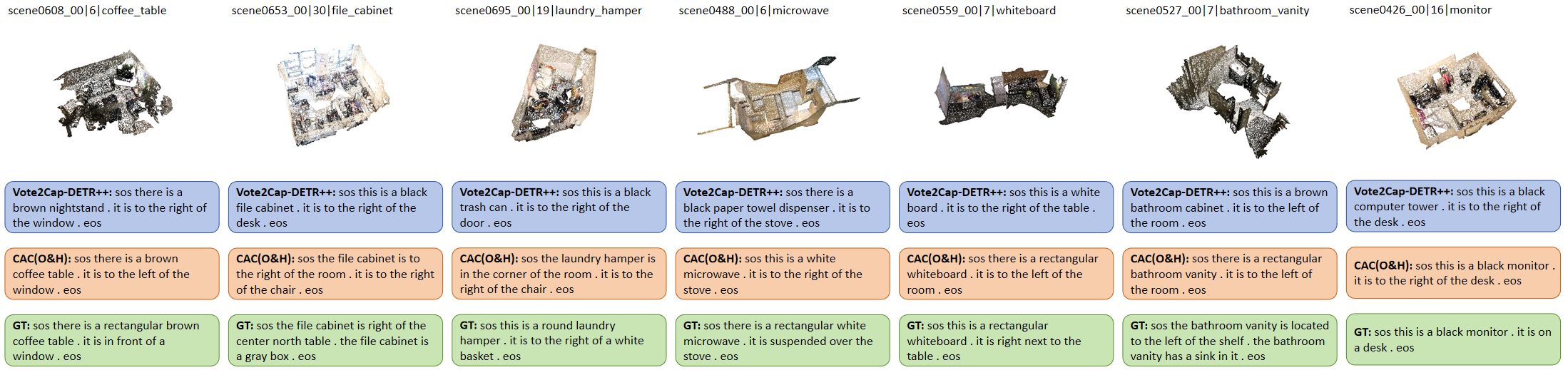}
\caption{
Supplementary qualitative results on the ScanRefer benchmark.
}
\label{fig:appendix_zhiliang} 
\end{figure*}
\subsubsection{Selective Oblique Manifold Projection}
In processing sparse point clouds, we leverage Oblique Manifold advantages to achieve significant performance gains. Our approach specifically maps only the encoder to Oblique Manifold, contrasting this configuration with alternative methodologies. As Table~\ref{tab:cmpEncDec} details, ablation studies systematically evaluate separate encoder and decoder mappings to Oblique Manifold, while further comparing self-attention versus cross-attention mechanisms within the decoder. To demonstrate Oblique Manifold superiority, we additionally map the encoder to Hyperbolic Manifold, providing direct comparison against our proposed framework.

\subsubsection{Hyperparameter sweeps and robustness}

We clarify that our training settings match the baseline. As space limited, the revised manuscript will include more ablation study across curvature values ($c \in [-2.0, -0.5]$) and temperature settings ($\tau \in [0.1, 2.0]$). We find curvature boosts performance, and our temperature modulates the sharpness of attention weights, our setting is a stable choice and there are still rooms for improvement.

\subsection{Qualitative Analysis}

While the baseline Vote2Cap-DETR++ is prone to object misclassification, our proposed CAC(\ensuremath{\mathcal{O}\&\mathbb{H}}) precisely captures both fine-grained object semantics and complex spatial relations. Consequently, our approach generates highly faithful linguistic descriptions that closely align with the ground truth annotations. Further detailed qualitative comparisons are illustrated in Figure ~\ref{fig:appendix_zhiliang}.

\section{Limitations}
Although the proposed Context-Aware Calibration (CAC) framework substantially mitigates semantic confusion and spatial hallucinations in 3D dense captioning, it is not without limitations. First, the integration of multi-level discriminative rewards via self-critical sequence training (SCST) inevitably introduces additional computational overhead and extends convergence time during the training phase. Second, as our framework heavily relies on scene graphs and 3D proposal generation paradigms, the quality of the generated captions remains inherently bounded by the efficacy of the upstream 3D object detector. Instances involving severe occlusions or highly sparse point clouds may still precipitate cascaded errors. Finally, our current empirical evaluations are predominantly centered on indoor scene datasets. Extending and adapting these context-aware calibration mechanisms to large-scale, unstructured outdoor environments presents a compelling avenue for future research.

\begin{table}[htbp]
    \centering
    \footnotesize
    \setlength{\tabcolsep}{5pt}
    \resizebox{\linewidth}{!}{ 
    \begin{tabular}{l|cccc}
    \hline
    Pretrain Configuration & mAP0.25 &  mAP0.50 & AR0.25 & AR0.50\\ \hline
    $\tau_{obl}$=1 (Epoch 119) & 61.04 & 38.05 & 86.23 & 61.57 \\
    $\tau_{obl}$=0.5 (Epoch 119) & 59.61 & 36.43 & 83.68 & 57.60 \\
    $\tau_{obl}$=2 (Epoch 119) & 61.19 & 37.77 & 85.91 & 59.70 \\
    \hline
    MLE Configuration with $\tau_{obl}$=1&  C@0.5$\uparrow$ & B-4@0.5$\uparrow$ & M@0.5$\uparrow$ & R@0.5$\uparrow$\\ \hline
    $\tau_{obl}$=1, $\tau_{lor}$=1, $c$=1 & 50.03 & 28.88 & 26.12 & 55.93 \\
    $\tau_{obl}$=1, $\tau_{lor}$=2, $c$=1 & 50.22 & 28.67 & 26.22 & 55.80 \\
    $\tau_{obl}$=1, $\tau_{lor}$=0.1, $c$=1 & 50.61 & 29.19 & 26.10 & 56.03 \\
    $\tau_{obl}$=1, $\tau_{lor}$=0.1, $c$=2
    & 50.87 & 29.28 & 26.29 & 56.11 \\
    $\tau_{obl}$=1, $\tau_{lor}$=0.1, $c$=0.5
    & 49.90 & 28.73 & 26.18 & 56.05 \\
    \hline
    \end{tabular}
    }
    \caption{
        Hyperparameter Sensitivity Analysis.
    }
    \label{tab:hyperparameter_sensitivity}
\end{table}


{
    \small
    \bibliographystyle{ieeenat_fullname}
    \bibliography{main}
}


%% file: sec/0_abstract.tex
\begin{abstract}
Accurate 3D scene description is fundamental to robotic navigation and augmented reality, yet current dense captioning methods face significant limitations in processing sparse point cloud data.
Existing approaches that apply Euclidean embedding spaces struggle to simultaneously preserve fine-grained local geometric details and model exponentially growing global semantic hierarchies, leading to either inaccurate localization or disjointed, shallow scene descriptions.
In this work, we propose a novel \textbf{\textsc{Curvature-Aware Captioning}} framework, integrating novel non-Euclidean geodesic attention mechanisms, to resolve the localization-contextualization conflict. 
Specifically, self-attention within Oblique space enforces dimensional homogeneity while establishing long-range dependencies. 
Bidirectional geodesic cross-attention within Lorentz space models hierarchical semantic relationships across scene instances, enabling simultaneous precision in object localization and coherence in scene descriptions.
Theoretical analysis confirms that the curvature complementarity between the Oblique manifold and Lorentz hyperboloid resolves the Euclidean-hyperbolic conflict, ensuring feature stability via isotropic optimization while preserving inherent hierarchical relationships.
Extensive experiments on ScanRefer and Nr3D benchmarks demonstrate state-of-the-art performance, with significant gains in both localization accuracy and descriptive richness.
\end{abstract}

%% file: sec/1_intro.tex
\section{Introduction}
3D dense captioning, which requires simultaneous object localization and descriptive caption generation within 3D scenes (e.g., point clouds)~\cite{chen2023executing, lin2021learning}, has seen significant methodological evolution. Early approaches predominantly adopted a ``detect-then-describe" pipeline~\cite{cai20223djcg, chen2022d, chen2021scan2cap, jiao2022more, wang2022spatiality}, suffering from cumulative errors. End-to-end set prediction frameworks address this by decoupling localization and caption decoding, exemplified by methods like \textit{Vote2Cap-DETR++}~\cite{chen2024vote2cap}.

Recent end-to-end transformer models like SpaCap3D~\cite{wang2022spatiality}, UniT3D~\cite{chen2023unit3d}, Vote2Cap-DETR++~\cite{chen2024vote2cap} and BiCA~\cite{kim2024bi} have achieved state-of-the-art (SOTA) performance in 3D dense captioning by leveraging unified attention mechanisms to integrate multi-modal features. These frameworks dynamically align visual cues with linguistic context through cross-modal attention, effectively resolving spatial-semantic dependencies while maintaining geometric consistency across scales. 
Notably, the recently proposed BiCA~\cite{kim2024bi} method introduces a \textit{bidirectional attention mechanism} that excels at integrating relationships between instance features and contextual features.

However, while capturing both local object features and global scene context, existing methods still struggle with \textbf{geometric conflicts} between fine-grained spatial localization and hierarchical semantic reasoning. 
Specifically, local object cues (e.g., surface geometry) demand Euclidean-like flat metrics, while global context (e.g., object hierarchies) inherently aligns with hyperbolic curvature for representing exponentially growing semantic distances~\cite{bronstein2017geometric}. Prior works in non-Euclidean deep learning~\cite{lai2025hunyuan3d} and hierarchical representation~\cite{zhang2021hype, zhao2025assembler} suggest that modeling such structural biases requires dedicated geometric spaces:

\begin{itemize}
    \item\textbf{\textsc{Oblique Manifold}} (\ensuremath{\mathcal{O}^{d\times k}}): 
    Ensures stability via $\left\|\mathbf{W}_{:,i}\right\|_{2}=1,\ \forall i\in\{1,\ldots, k\}$
    \item\textbf{\textsc{Lorentz Space}} (\ensuremath{\mathbb{H}^n_\mathscr{L}}): 
    Models hierarchies via \ensuremath{\mathscr{G}_{\mathbb{H}_\mathscr{L}}(\mathbf{u},\mathbf{v}) = \operatorname{arcosh}\left(-\langle\mathbf{u}, \mathbf{v}\rangle_{\mathscr{L}}\right)}
\end{itemize}
\noindent where column normalization in $\mathcal{O}$ transforms optimization landscapes~\cite{qi2021transductive, absil2006joint}, and $\mathscr{G}_{\mathbb{H}_\mathscr{L}}$ denotes the Lorentzian distance metric~\cite{hedicke2024lorentzian}.

Inspired by these geometric advantages and building upon BiCA, we propose \textbf{\textsc{Curvature-Aware Captioning} (CAC)}, a novel framework that resolves attentional conflicts through \textit{stage-specific manifold projection}. 

The Oblique Manifold projection enables near-linear gradient descent paths, accelerating convergence while preserving geometric integrity, significantly stabilizing bounding box regression and enhancing localization precision. Complementarily, Lorentz Space captures exponentially growing semantic hierarchies through hyperbolic distances, dynamically modeling compositional relationships between objects at varying scales. These geometric transformations collectively resolve spatial-semantic conflicts while maintaining coherence between object localization and descriptive captioning.
\noindent Our main contributions are summarized as follows:
\begin{itemize}
    \item \textbf{Encoder-Oblique Projection}: In the encoding and object localization stage, 
    we utilize geodesic attention in Oblique manifold, promoting optimization stability.
    \item \textbf{Decoder-Lorentz Contextualization}: During decoding (caption generation), 
    we extend bidirectional contextual attention in Lorentz Space, where hyperbolic geodesics capture hierarchical relations.
    \item \textbf{State-of-the-art performance} on ScanRefer and Nr3D benchmarks, achieving $\uparrow$\textbf{2.7\% CIDEr@0.5} on ScanRefer and $\uparrow$\textbf{4.6\% CIDEr@0.5} on Nr3D the highest reported score to date. Our work pioneers stage-specific manifold learning for 3D vision-language tasks, achieving geometrically consistent scene understanding.
\end{itemize}

\begin{figure*}
\centering
\includegraphics[width=1\linewidth]{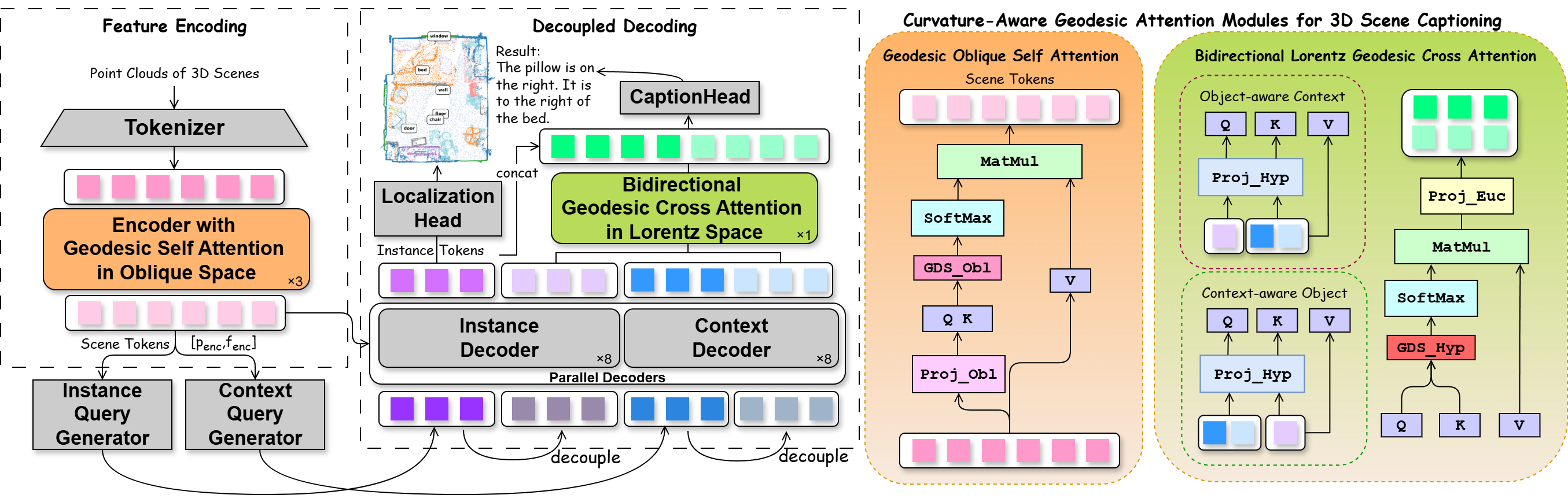}
\caption{
Left: Overview of the proposed Curvature-Aware Captioning (CAC) framework. 
The processing pipeline is mapped onto non-Euclidean manifolds, with attention computed via geodesic distance metrics. 
Right: The Geodesic Oblique Self Attention and Bidirectional Lorentz Geodesic Attention.
}
\label{fig:liuchengtu} 
\end{figure*}

%% file: sec/2_relat.tex
\section{Related Work}

\subsection{3D Dense Captioning}

3D dense captioning, an evolving task demanding precise localization and descriptive generation of objects in complex 3D scenes~\cite{dai2017scannet}, has progressed through diverse methodological innovations, starting with graph-based techniques that treat detector proposals as nodes in k-nearest neighbor graphs for feature extraction~\cite{mao2023complete,jiao2022more}. Subsequent transformer-based methods replaced explicit graph operations with attention mechanisms to better model spatial relationships~\cite{cai20223djcg,wang2022spatiality,jin2023context}, while recent efforts explore multi-task synergies between dense captioning, 3D visual grounding~\cite{chen2020scanrefer}, and 3D question answering~\cite{azuma2022scanqa,ye20223d}. Standard architectures comprise scene encoders employing detectors such as~\cite{misra2021end,qi2017pointnet++,jiang2020pointgroup} for object-level feature extraction, relational modules modeling inter-object connections through graph-based, transformer-based, or knowledge distillation approaches~\cite{chen2023end,chen2024vote2cap}, and feature decoders generating bounding boxes and captions using GRU-based or transformer-based implementations~\cite{yuan2022x}.

\subsection{Embedding on Oblique Manifold}

The Oblique Manifold~\cite{qi2021transductive} is an embedded submanifold defined by column-wise unit Euclidean norm constraints. This space finds critical application in independent component analysis~\cite{absil2006joint}, where its intrinsic normalization property (analogous to $L_2$-normalization) facilitates uniform scaling in gradient-based optimization, enhancing convergence efficiency and model accuracy~\cite{guo2021sparse}. Geometrically, it transforms elongated contour profiles into near-spherical geometries with isotropic curvature, mitigating directional bias to accelerate convergence while reducing susceptibility to suboptimal solutions. Formal definitions appear in the Appendix.

\subsection{Hyperbolic Geometry for Point Clouds}
\label{subsec:hyperbolic_pointclouds}

Recent advances leverage hyperbolic embeddings to exploit the hierarchical nature of 3D point clouds, capitalizing on hyperbolic geometry's capacity for efficient representation of high-dimensional data in low-dimensional spaces~\cite{cetin2022hyperbolic}. Beyond foundational hierarchy modeling with graph-based approaches using InfoNCE~\cite{nickel2017poincare} and entailment losses~\cite{ganea2018hyperbolic}, domain-specific adaptations have gained prominence: Hyperbolic ProtoNet facilitates few-shot visual classification~\cite{khrulkov2020hyperbolic}, while HypRe~\cite{montanaro2022rethinking} pioneers explicit regularization for part-whole relationships. Extensions enhance 3D processing through attentional mechanisms (PHGT~\cite{liu2024application}), multispace fusion (HypLiLoc~\cite{wang2023hypliloc}), and Riemannian-geometry modules improving point cloud matching (HECPG~\cite{xie2024hecpg}). Complementing these, Hypformer~\cite{yang2024hypformer} demonstrates how Lorentzian temporal dimensions intrinsically induce hierarchical structures to capture inherent relationships within point clouds. 

\begin{figure}
\centering
\includegraphics[width=1\linewidth]{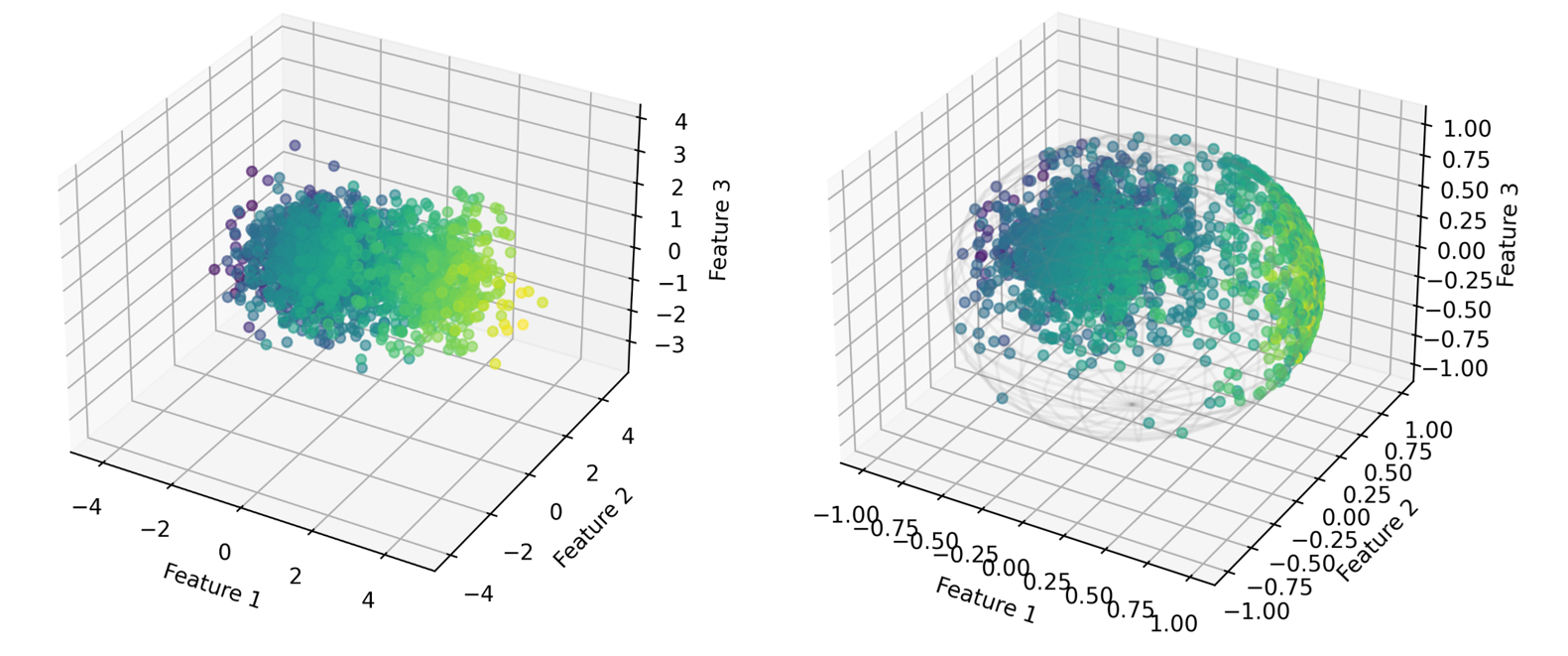}
\caption{
3D point cloud distribution contrast: Euclidean space (left) versus Oblique manifold projection (right) with column-wise unit Euclidean norm constraints.
}
\label{fig:oblpro} 
\end{figure}

%% file: sec/3_preliminary.tex
\section{Preliminary}
This section introduces core mathematical foundations for our architecture: the Oblique Manifold and Lorentz Space. These frameworks overcome Euclidean representation limitations in 3D geometric understanding.

\subsection{Oblique Manifold}

The Oblique Manifold ($\mathcal{OM}$) constitutes a Riemannian submanifold of Euclidean space defined by column-wise unit Euclidean norm constraints. This geometric structure underpins independent component analysis~\cite{absil2006joint}, where intrinsic orthonormality facilitates efficient feature separation through constrained optimization.Formally, $ \mathcal{OM}(n,g) $ is defined as:
\begin{equation} \label{eq:om_def}
\mathcal{OM}(n,g) = \left\{ \bm{P} \in \mathbb{R}^{n \times g} : \operatorname{diag} \left( \bm{P}^{T} \bm{P} \right) = \bm{I}_{g} \right\},
\end{equation}
where $\operatorname{diag}(\cdot)$ denotes the diagonal matrix, $n$ represents the feature dimension of each patch vector.

Projecting point cloud features from Euclidean space to $\mathcal{OM}$ preserves relative geometric relationships~\cite{li2022geodesic}, satisfying attention mechanisms' fundamental requirement to capture relational dependencies.

\begin{equation} \label{eq:ob_proj}
\bm{P} := \operatorname{Proj}(\bm{P}) = \operatorname{Cat}\left( \left\{ \frac{p_{i}}{\left\| p_{i} \right\|} \right\}_{i=1}^{g} \right),
\end{equation}
where $\bm{P} = \{ p_1, p_2, \cdots, p_g \}$ is the point cloud patch matrix comprising $g$ patches, each $p_i$ representing a feature vector of the $i$-th patch. Here $\operatorname{Cat}(\cdot)$ denotes the concatenation function, $\|\cdot\|$ denotes the squared Frobenius norm in the ambient space, as shown in Figure~\ref{fig:oblpro}.

After projection, the geodesic distance of the input point pair $\{\bm{Q}, \bm{K}\}$ on $\mathcal{O}\mathcal{M}$ can be calculated as:
\begin{equation} \label{eq:ob_dist}
\operatorname{dist}(\bm{Q}, \bm{K}) = \sqrt{\sum_{i=1}^{n} \arccos^{2} \left( \operatorname{diag} \left( \bm{Q}^{T} \bm{K} \right) \right)_{i}}.
\end{equation}




\subsection{Lorentz Space}

Hyperbolic space offers a fundamental non-Euclidean framework for hierarchical representation through constant negative curvature. We employ the Lorentz hyperboloid model~\cite{Ramasinghe2024} to avoid numerical instabilities from exponential volume growth during training, replacing alternatives like Poincaré ball, Klein, and Poincaré half-space models~\cite{shimizuhyperbolic2021} projecting onto space-like hyperplanes.
\begin{figure}
\centering
\includegraphics[width=1\linewidth]{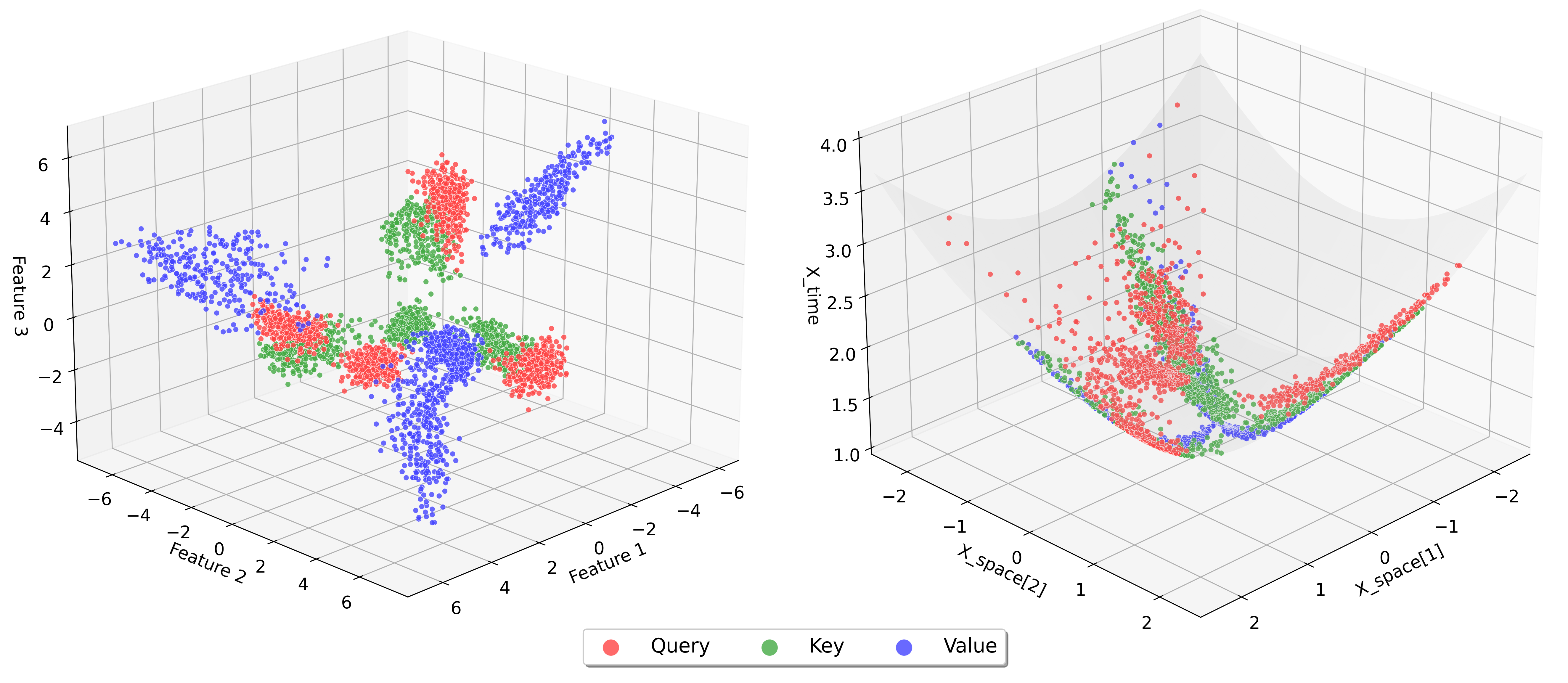}
\caption{
Comparison of 3D point distributions: Euclidean Cartesian space (left) versus Lorentz Space embeddings (right). Hyperbolic projections via exponential mapping reveal hierarchical clustering patterns, with semitransparent boundaries indicating the manifold structure.
}
\label{fig:enhanced_lorentz_contrast} 
\end{figure}

Hyperbolic space is formalized through the \textbf{Lorentz hyperboloid model} defined as:
\begin{equation}
\mathbb{L}^{n}=\left\{\mathbf{x}\in\mathbb{R}^{n+1}:\langle\mathbf{x},\mathbf{x}\rangle_{\mathbb{L}}=-1/c\right\}(c>0),
\end{equation}
where $c > 0$ determines curvature $-c$. Vectors decompose as $\mathbf{x} = [\mathbf{x}_{\text{space}}, \mathbf{x}_{\text{time}}]$ with $\mathbf{x}_{\text{time}} = \sqrt{1/c + \lVert \mathbf{x}_{\text{space}} \rVert^2}$. The Lorentzian inner product follows:
\begin{equation}
\langle\mathbf{x},\mathbf{y}\rangle_{\mathbb{L}}=\mathbf{x}_{\text{space}} \cdot \mathbf{y}_{\text{space}}-\mathbf{x}_{\text{time}} \cdot\mathbf{y}_{\text{time}}.
\end{equation}
To mitigate numerical instability during training~\cite{kim2024hype}, we exclusively map through the origin $\mathbf{O} = [\mathbf{0}, 1/\sqrt{c}]$. 
The exponential map $\exp_{\mathbf{O}}: \mathcal{T}_\mathbf{O}\mathbb{L}^n \to \mathbb{L}^n$ projects Euclidean tangent vectors $\mathbf{u} = [\mathbf{u}_{\text{enc}}, 0]$ to the manifold:
\begin{equation} \label{lorentz_prj}
\mathbf{x}_{\text{space}} = \frac{\sinh(\sqrt{c} \lVert \mathbf{u}_{\text{enc}} \rVert)}{\sqrt{c} \lVert \mathbf{u}_{\text{enc}} \rVert} \mathbf{u}_{\text{enc}},
\end{equation}
where $\mathbf{u}_{\text{enc}} \in \mathbb{R}^n$ represents the Euclidean embedding component of the tangent vector at the origin $\mathbf{O}$.
Conversely, the logarithmic map $\log_{\mathbf{O}}: \mathbb{L}^n \to \mathcal{T}_\mathbf{O}\mathbb{L}^n$ projects manifold points to Euclidean space:
\begin{equation} \label{log_map}
\mathbf{u}_{\text{enc}} = \frac{\operatorname{arcosh}(-\langle \mathbf{x}, \mathbf{O} \rangle_{\mathbb{L}})}{\sqrt{\langle \mathbf{x}, \mathbf{O} \rangle_{\mathbb{L}}^2 - 1}} \mathbf{x}_{\text{space}},
\end{equation}
where $\mathbf{u}_{\text{enc}} = \alpha f(T)$ and $\alpha$ (initialized as $1/\sqrt{n}$) is learned in logarithmic space to prevent embedding collapse and numerical overflow~\cite{DBLP:conf/icml/DesaiNR0V23}. Analogous scaling applies to image embeddings $\mathbf{v}_{\text{enc}} = \alpha_{\text{img}} g(I)$. \textit{Geodesics} in $\mathbb{L}^n$ correspond to hyperplane-origin intersections with distance metric:
\begin{equation} \label{hyper_dist}
\mathscr{G}_{\mathbb{H}_\mathscr{L}}(\mathbf{x},\mathbf{y}) = \frac{1}{\sqrt{c}} \cosh^{-1}\left( -c \langle\mathbf{x},\mathbf{y}\rangle_{\mathbb{L}} \right).
\end{equation} 
This formulation intrinsically captures hierarchical relationships while maintaining geometric consistency across spatial scales, enabling robust optimization in negatively curved spaces.

%% file: sec/4_method.tex
\section{Methodology}

This section introduces \textbf{CAC}, a geometrically unified framework addressing spatial and hierarchical representation limitations in 3D dense captioning. We demonstrate curvature complementarity between Oblique and Lorentz spaces resolves Euclidean-hyperbolic conflicts (Appendix derivations). The input 3D point clouds $\text{PC} = [p_{\text{in}}; f_{\text{in}}] \in \mathbb{R}^{40,000 \times (3 + F)}$ comprise $40,000$ points with positional coordinates (xyz) and $F$-dimensional features.

\subsection{Overall Architecture}

Current state-of-the-art 3D dense captioning methods inadequately leverage spatial and hierarchical structures, neglecting inherent geometric priors in point clouds. We resolve this limitation through \textbf{CAC}, extending Vote2Cap-DETR++ to migrate data into geometric-prior-optimized space, with ablation studies confirming enhanced feature extraction. 

Recognizing hyperbolic space's innate hierarchical representation capacity, we introduce hyperbolic multi-head attention that transitions both Object-aware Contexts and Context-aware Objects modules into hyperbolic space, as shown in Figure~\ref{fig:liuchengtu}. This enables explicit hierarchical modeling while incorporating geodesic distance mechanisms that amplify inherent geometric priors. Our framework employs geometrically consistent geodesic attention operators, simultaneously leveraging 3D spatial priors and hierarchical structures.

\subsection{Encoder on Oblique Manifold}

In this work, in order to advance sparse 3D point cloud embedding, we replace the 3DETR encoder's attention module ~\cite{misra2021end} by our  \textit{Geodesic Oblique Self Attention} mechanism, as shown in Figure~\ref{fig:liuchengtu}.
Specifically, following Vote2Cap-DETR++'s preprocess pipeline, PointNet++'s set-abstraction layer~\cite{qi2017pointnet++} tokenizes $\text{PC}$ into coordinates $p_{\text{abs}} \in \mathbb{R}^{2,048 \times 3}$ and feature representations $f_{\text{abs}} \in \mathbb{R}^{2,048 \times 256}$, where $2,048$ denotes the sampled point count and $256$ denotes the token embedding dimension.
Subsequently processed by the geometrically enhanced 3DETR encoder, this framework outputs encoded scene tokens $p_{\text{enc}} \in \mathbb{R}^{2,048 \times 3}$ and $f_{\text{enc}} \in \mathbb{R}^{2,048 \times 256}$.
%
%
These tokens are subsequently processed by the geometrically enhanced 3DETR encoder and downsampled to yield the encoded scene tokens $\mathbf{p}_{\text{enc}} \in \mathbb{R}^{1,024 \times 3}$ and $\mathbf{f}_{\text{enc}} \in \mathbb{R}^{1,024 \times 256}$.

The \textit{Geodesic Oblique Self Attention} mechanism imposes Oblique Manifold constraints to ensure isotropic attention distribution~\cite{long2023molecular}. 
The mechanism projects embedded features ($\textbf{Q},\textbf{K},\textbf{V} \in \mathbb{R}^{2048 \times 256}$) onto the Oblique manifold using equation (\ref{eq:ob_proj}). This projection enables establishing spatial-geometric relationships between queries and keys through pairwise geodesic distances $D \in \mathbb{R}^{2048 \times 2048}$ computed via equation (\ref{eq:ob_dist}), which intrinsically respect the manifold curvature. 
To ensure numerical stability during the computation of these large-scale pairwise distances, we employ a clipping operation with $\epsilon = 10^{-4}$ to constrain the inputs within the valid domain $[-1+\epsilon, 1-\epsilon]$, preventing potential numerical instabilities.
Differentiable aggregation modulates softmax-weighted feature fusion using these geodesic similarity scores, boosting object recognition accuracy by capturing intrinsic spatial structures and contextual relationships in 3D point clouds, please refer to Algorithm~\ref{alg:Oblique_geodesic}.

\begin{algorithm}
\caption{Geodesic Oblique Self Attention}
\label{alg:Oblique_geodesic}
\begin{algorithmic}[1]
\REQUIRE{$X \in \mathbb{R}^{n \times d}$, $Pos \in \mathbb{R}^{n \times 3}$}
\ENSURE{Geodesic attention output $\hat{v} \in \mathbb{R}^{n \times d}$}
\STATE $ q,k \gets \text{Proj}_{\mathcal{OBL}}(\text{Emb}(X, Pos))$ \COMMENT{Via equation(\ref{eq:ob_proj})}
\STATE $v \gets X$
\FOR{$i = 1$ to $n$}
    \FOR{$j = 1$ to $n$}
        \STATE $d_{ij} \gets GDS_{\mathcal{OBL}}(q_i, k_j)$ \COMMENT{Via equation(\ref{eq:ob_dist})}
    \ENDFOR
\ENDFOR
\STATE $\hat{v} \gets MatMul \langle \text{softmax}(-D),v \rangle$
\RETURN $\hat{v}$ \COMMENT{Geodesic attention output}
\end{algorithmic}
\end{algorithm}

\subsection{Vote Query}
Inspired by Vote2Cap-DETR++,our Vote Query Generator produces queriesthrough a spatial refinement process:  
\begin{equation}
[\Delta p_{\text{vote}}; \Delta f_{\text{vote}}] = \text{FFN}(f_{\text{enc}}),
\end{equation}
\begin{equation} \label{pvq}
(p_{vq},f_{vq}) = \text{SA}(p_{\text{enc}} + \Delta p_{\text{vote}}, f_{\text{enc}} + \Delta f_{\text{vote}}),
\end{equation}
where $\Delta p_{\text{vote}} \in \mathbb{R}^{1,024\times3}$ learns center-aligned spatial shifts by a \textbf{F}eed \textbf{F}orward \textbf{N}etwork(FFN), and 
$\text{SA}$ denotes the set-abstraction layer (radius=0.3, 16-point sampling) that aggregates features at refined coordinates, sampling $\text{npoint}=256$ points.

Crucially extending~\cite{chen2024vote2cap}, we employ a decoupled decoding structure that iteratively updates spatial locations across layers:
\begin{equation}
\begin{aligned}
f_{\text{query}}^{i} &= \text{Layer}_{i-1} \left( f_{\text{query}}^{i-1} + \text{PE} \left( p_{vq}^{i} \right) \right), \\
p_{vq}^{i} &= p_{vq}^{i-1} + \text{FFN} \left( f_{\text{query}}^{i} \right),
\end{aligned}
\end{equation}
where $\text{Layer}_{i-1}$ denotes the $i$-th decoder layer and $\text{PE}(\cdot)$ represents 3D Fourier positional encoding, mitigating cumulative errors inherent in detect-then-describe pipelines~\cite{zhang2022dino,zhu2020deformable}.

Both Instance and Context Query Generators employ iterative spatial refinement, with Context processing excluding $\Delta p_{\text{vote}}$ fusion during aggregation. 

\subsection{Decoder with Bidirectional Lorentz Attention}

Our decoder maintains Vote2Cap-DETR++'s decoupled architecture~\cite{chen2024vote2cap}, processing both encoder outputs and vote queries from the generator. This dual-decoder framework generates instance and context features, outputting $p_{\text{dec}} \in \mathbb{R}^{npoint \times 2 \times 3}$ and $f_{\text{dec}} \in \mathbb{R}^{npoint \times 2 \times 256}$ where $npoint=256$ denotes the number of sampled points during vote query processing and $2$ denotes the decoupled outputs. For caption generation, we utilize the first dimension of decoupled instance features $f_{\text{cap}} \in \mathbb{R}^{npoint \times 1 \times 256}$ as input to the caption task head to obtain object segmentation information.

As depicted in Figure~\ref{fig:liuchengtu}, we enhance cross-attention by constraining Key and Value matrices within hyperbolic geometry~\cite{alayrac2022flamingo}, replacing dot products with geodesic distances to leverage 3D hierarchical structures. Following feature extraction, hyperbolic bidirectional attention modules implement distinct configurations: Object-aware Context (OAC) employs instance features as \textbf{Q} with context features as \textbf{K}/\textbf{V}, while Context-aware Object (CAO) utilizes context features as \textbf{Q} with instance features as \textbf{K}/\textbf{V}, with ablation studies confirming significant performance gains.

For bidirectional attention inputs, we utilize the second dimension of decoupled instance features $f_{\text{ins}} \in \mathbb{R}^{\text{npoint} \times 1 \times 256}$ and both dimensions of context features $f_{\text{ctx}} \in \mathbb{R}^{\text{npoint} \times 2 \times 256}$. 
%
Features undergo Lorentz projection into hyperbolic space (Equation~\ref{lorentz_prj}, Figure~\ref{fig:enhanced_lorentz_contrast}), establishing geometric foundations for attention computation. Attention weights derive from pairwise geodesic distances (Equation~\ref{hyper_dist}) that intrinsically capture hyperbolic curvature relationships between queries and keys, replacing conventional dot products. To maintain numerical stability during this computation, we implement a clipping operation with $\epsilon = 10^{-15}$ to constrain inputs within the valid domain $[1+\epsilon, \infty]$. This ensures the argument of the $\operatorname{arcosh}(\cdot)$ function remains strictly greater than or equal to unity, thereby preventing numerical instabilities near boundary conditions. These stabilized distances subsequently undergo exponential mapping and temperature-scaled softmax normalization to produce curvature-aware weights. The architecture aggregates contextual information through geodesic-weighted combinations before Euclidean reprojection (Equation~\ref{log_map}), completing the attention cycle as detailed in Algorithm~\ref{alg:hyperbolic_geodesic}. Final processing applies mean pooling to Context-aware Object outputs, yielding feature representations $f_{\text{ins}}$, $f_{\text{oac}}$, $f_{\text{cao}} \in \mathbb{R}^{n_{\text{point}} \times 1 \times 256}$ that are concatenated as input to the captioning task head.

\begin{algorithm}
\caption{Bidirectional Lorentz Geodesic Attention}
\label{alg:hyperbolic_geodesic}
\begin{algorithmic}[1]
\REQUIRE{$q, k, v \in \mathbb{R}^{n \times b \times d}$}
\ENSURE{Geodesic attention output $\hat{v} \in \mathbb{R}^{n \times b \times d}$}

\STATE $q_\mathbb{H},k_\mathbb{H} \gets \text{Proj}_{\mathcal{HYP}}(q,k)$ 
\COMMENT{Via equation~\eqref{lorentz_prj}}
\FOR{$i = 1$ to $n$}
    \FOR{$j = 1$ to $n$}
        \STATE $d_{ij} \gets GDS_{\mathcal{HYP}}(q_{\mathbb{H}i}, k_{\mathbb{H}j})$ \COMMENT{Via equation~\eqref{hyper_dist}}
    \ENDFOR
\ENDFOR
\STATE $A \gets \text{softmax}(\exp(-D/\tau))$ \COMMENT{$\tau$: temperature}

\STATE $\hat{v} \gets MatMul \langle A,v \rangle$
\RETURN $\hat{v}$
\end{algorithmic}
\end{algorithm}

\input{tab/1_all}

\subsection{Training CAC}

Our loss builds upon Vote2Cap-DETR++~\cite{chen2024vote2cap}, adopting its \textit{decoupled localization-describing paradigm}. The objective integrates four weighted components: $\mathcal{L}_{\text{vq}}$ supervising point shifts to object centers, $\mathcal{L}_{\text{det}}$ refining proposal assignment via Hungarian algorithm, $\mathcal{L}_{\text{cap}}$ implementing dual training (MLE+SCST), and $\mathcal{L}_{\text{qr}}$ enabling iterative query localization across decoder layers.

\subsubsection*{Detection Loss}
Employing 3DETR's Hungarian algorithm~\cite{misra2021end} for proposal-ground truth matching, we emphasize 3D GIoU loss to enhance localization:
\begin{equation}
\mathcal{L}_{\text{set}}=\alpha_{1}\mathcal{L}_{\text{giou}}+\alpha_{2}\mathcal{L}_{\text{cls}}+\alpha_{3}\mathcal{L}_{\text{center-reg}}+\alpha_{4}\mathcal{L}_{\text{size-reg}},
\label{eq:det_loss}
\end{equation}
with heuristic weights $\alpha_1\!=\!10$, $\alpha_2\!=\!1$, $\alpha_3\!=\!5$, $\alpha_4\!=\!1$. This loss aggregates across all $n_{\text{dec-layer}}$ decoder layer.

\subsubsection*{Caption Loss}
Our caption head integrates Maximum Likelihood Estimation (MLE) and \textit{Self-Critical Sequence Training} (SCST)~\cite{rennie2017self}. The MLE objective minimizes:
\begin{equation}
\mathcal{L}_{c_{i}}=-\sum_{t=1}^{T}\log\hat{P}\left(c_{i}^{t+1}\mid\mathcal{V};c_{i}^{[1:t]}\right),
\label{eq:mle_loss}
\end{equation}
while SCST fine-tuning employs beam search (size $k$) and greedy decoding:
\begin{equation}
\mathcal{L}_{c_{i}}=-\sum_{i=1}^{k}\left(R(\hat{c}_{i})-R(\hat{g})\right)\cdot\frac{1}{|\hat{c}_{i}|}\log\hat{P}\left(\hat{c}_{i}\mid\mathcal{V}\right).
\label{eq:scst_loss}
\end{equation}
Here $R(\cdot)$ represents the CIDEr reward metric~\cite{vedantam2015cider}, with length normalization $|\hat{c}_{i}|$ ensuring balanced caption weighting, where $\hat{g}$ denotes the greedily decoded baseline caption and $\hat{c}_{i}$ represents the $i$-th beam search candidate.

\subsubsection*{Set-to-Set Training and Integrated Objectives}
Our \textit{set-to-set strategy} randomly samples one caption per instance, assigns annotations via Hungarian matching, and averages batch caption losses $\mathcal{L}_{c_i}$ into $\mathcal{L}_{\text{cap}}$. The unified objectives are:
\begin{align}
\mathcal{L}_{\text{V1}} &= \beta_{1}\mathcal{L}_{\text{vq}} + \beta_{2}\sum_{i=1}^{n_{\text{dec-layer}}}\mathcal{L}_{\text{set}} + \beta_{3}\mathcal{L}_{\text{cap}} \label{eq:total_loss_v1}, \\
\mathcal{L}_{\text{V2}} &= \mathcal{L}_{\text{V1}} + \beta_{4}\sum_{i\in\delta}\mathcal{L}_{\text{qr}},
\label{eq:total_loss_v2}
\end{align}
with $\beta_1\!=\!\beta_4\!=\!10$, $\beta_2\!=\!1$, $\beta_3\!=\!5$ applied to decoder layers $\delta$ performing spatial refinement.

%% file: tab/1_all.tex
\begin{table*}[htbp]
    
    \centering
    \footnotesize 
    \begin{tabular}{ccccccccccccccccc}
    \toprule
    \multirow{2}{*}{Method} & \multirow{2}{*}{$\mathcal{L}_{des}$}
                            &                                      &  & \multicolumn{4}{c}{IoU = 0.25}                          &  & \multicolumn{4}{c}{IoU = 0.50}                          &                                \\ 
                            \cline{4-7} \cline{9-12} \cline{14-17}
                            &                                      &  & C$\uparrow$ & B-4$\uparrow$ & M$\uparrow$ & R$\uparrow$ &  & C$\uparrow$ & B-4$\uparrow$ & M$\uparrow$ & R$\uparrow$ &   \\ \hline
    Scan2Cap~\cite{chen2021scan2cap}   & \multirow{12}{*}{MLE} &  & 53.73       & 34.25         & 26.14       & 54.95       &  & 35.20       & 22.36         & 21.44       & 43.57       &       \\
    MORE~\cite{jiao2022more}           &                       &  & 58.89       & 35.41         & 26.36       & 55.41       &  & 38.98       & 23.01         & 21.65       & 44.33       &       \\
    SpaCap3d~\cite{wang2022spacap3d}   &                       &  & -           & -             & -           & -           &  & 42.76       & 25.38         & 22.84       & 45.66       &     \\
    Contextual~\cite{zhong2022contextual} &                    &  & -           & -             & -           & -           &  & 42.77       & 23.60         & 22.05       & 45.13       &   \\
    3DJCG~\cite{cai20223djcg}          &                       &  & 60.86       & 39.67         & 27.45       & 59.02       &  & 47.68       & 31.53         & 24.28       & 51.80       &        \\
    %
    %
    3D-VLP$^{*}$~\cite{jin2023context} &                       &  & 64.09        & 39.84        & 27.65       & 58.78       &  & 50.02       & 31.87         & 24.53       & 51.17       &         \\
    Vote2Cap-DETR~\cite{chen2023end} &                                      
        &  & 71.45   & 39.34   & 28.25   & 59.33
        &  & 61.81   & 34.46   & 26.22   & 54.40\\

    Vote2Cap-DETR++~\cite{chen2024vote2cap}          &                                      
        &  & 76.36   & 41.37   & 28.70   & 60.00
        &  & 67.58   & 37.05   & 26.89   & 55.64\\


    \textbf{Vote2Cap-DETR++$^{R}$}          &                                      
        &  & 74.59   & 41.35   & 28.60   & 59.79
        &  & 66.06   & 36.93   & 26.76   & 55.39   \\

    \cline{1-1}
    \multicolumn{1}{l}{\textbf{\textit{Ours:}}} \\
    CAC(\ensuremath{\mathcal{O}})&                                      
        &  & 75.56   & 40.34   & 28.55   & 59.53     
        &  & 68.07   & 36.53   & 26.72   & 55.08\\     

    CAC(\ensuremath{\mathcal{O}\&\mathbb{H}})&                                      
        &  & \textbf{78.67}   & \textbf{42.11}   & \textbf{28.79}   & \textbf{60.22}
        &  & \textbf{69.92}   & \textbf{37.67}   & \textbf{26.89}   & \textbf{55.62}\\     

    \hline
    $\chi$-Trans2Cap~\cite{yuan2022x}        & \multirow{8}{*}{SCST}                &  & 58.81       & 34.17         & 25.81       & 54.10       &  & 41.52       & 23.83         & 21.90       & 44.97       &      \\
    Contextual~\cite{zhong2022contextual} &                                      &  & -       & -        & -       & -          &  & 50.29       & 25.64         & 22.57       & 44.71       & \\
    %
    %
    Vote2Cap-DETR~\cite{chen2023end} &                          
        &  & 84.15   & 42.51   & 28.47   & 59.26     
        &  & 73.77   & 38.21   & 26.64   & 54.71\\
    Vote2Cap-DETR++~\cite{chen2024vote2cap} &                                      
        &  & 88.28   & 44.07   & 28.75   & 59.89
        &  & 78.16   & 39.72   & 26.94   & 55.52\\
    \textbf{Vote2Cap-DETR++$^{R}$}          &                                      
        &  & 87.52   & 44.00   & 28.71   & 59.59 
        &  & 77.65   & 39.59   & 26.88   & 55.21 \\
    \cline{1-1}
    \multicolumn{1}{l}{\textbf{\textit{Ours:}}} \\
    CAC(\ensuremath{\mathcal{O}})&                          
        &  & 89.51   & 43.47   & 28.76   & 59.51    
        &  & 79.09   & 38.96   & 26.85   & 54.95  \\        
    CAC(\ensuremath{\mathcal{O}\&\mathbb{H}})&                                      
        &  & \textbf{90.16}   & \textbf{44.07}   & \textbf{28.80}   & \textbf{60.29}
        &  & \textbf{80.35}   & \textbf{39.95}   & \textbf{26.94}   & \textbf{55.66}\\
    \bottomrule
    \end{tabular}
    \caption{
        Quantitative comparisons on ScanRefer~\cite{chen2020scanrefer} validation set following Scan2Cap~\cite{chen2021scan2cap}'s protocol. Superscript $^{R}$ indicates reproduced results. Our framework establishes new state-of-the-art performance across caption supervision paradigms.
    }
    \label{tab:scanrefer}
\end{table*}

%% file: sec/5_experiment.tex
\section{Experiments}






\textbf{Datasets.}~ 
This study employs two benchmark datasets for 3D dense captioning: ScanRefer~\cite{zhenyu2019scanrefer} (36,665 descriptions across 7,875 objects in 562 scenes) and Nr3D~\cite{achlioptas2020referit3d} (32,919 descriptions for 4,664 objects in 511 scenes), both providing human-annotated scene and object descriptions. Training data originates from ScanNet's 1,201 distinct scenes. Evaluation utilizes 9,508 ScanRefer descriptions (2,068 objects in 141 scenes) and 8,584 Nr3D descriptions (1,214 objects in 130 scenes), all drawn from ScanNet's 312-scene validation set.


\textbf{Evaluation Metrics.}
To ensure evaluation consistency with prior works~\cite{cai20223djcg, chen2024vote2cap}, we apply Non-Maximum Suppression (NMS)~\cite{neubeck2006efficient} to predictions despite our method's inherent robustness. Post-NMS, we establish ground-truth correspondences by selecting highest-IoU proposals for each annotated instance $(b_i, C_i)$. Performance is evaluated using the \textbf{m@kIoU metric}~\cite{chen2021scan2cap}:
\begin{equation}
m@kIoU = \frac{1}{N} \sum_{i=1}^N m(\hat{c}_i, C_i) \cdot \mathbb{I}\{ IoU(b_i, \hat{b}_i) \geq k \},
\label{eq:mkiou}
\end{equation}
combining \textbf{localization accuracy} (IoU $\geq k$) and \textbf{caption quality} (metric $m$). Here $N$ enumerates instances, $m$ denotes caption metrics (CIDEr, METEOR, BLEU-4, ROUGE-L), and $\mathbb{I}\{\cdot\}$ is the indicator function.

\textbf{Implementation Details.}~
Following~\cite{chen2024vote2cap}, our training employs a three-stage pipeline. 
%
%
The initial phase pre-trains the model for 1,080 epochs on ScanNet~\cite{dai2017scannet} excluding caption modules, using AdamW optimization~\cite{loshchilov2017decoupled} with cosine annealing from $5\times10^{-4}$ to $10^{-6}$ learning rate, 0.1 weight decay, and gradient clipping at batch size 8. Subsequently, 720 epochs of joint training on ScanRefer and Nr3D apply cross-entropy loss with a batch size of 8, with fixed $10^{-6}$ detector learning rate while decaying the caption head from $10^{-4}$ to $10^{-6}$.
The final stage refines the caption head via SCST~\cite{rennie2017self} for 180 epochs at fixed $10^{-6}$ learning rate and batch size 2, freezing detector parameters. 
%
%
%
Implemented in PyTorch~\cite{paszke2019pytorch}, our framework exhibits distinct performance characteristics across its three training phases. 
Table~\ref{tab:training_configs} summarizes the basic configurations for different training phases.
During the pre-training stage, the average iteration time is 0.878s with a statistical sample size of 16,635 and memory consumption of approximately 18GB, achieving a per-sample computational complexity of 10.61 GFLOPs and a per-iteration complexity of 84.88 GFLOPs. The second phase attains an average iteration time of 1.173s over 5,163 samples, maintaining approximately 18GB memory usage, with a per-sample FLOPs of 11.16 GFLOPs and a per-iteration FLOPs of 89.30 GFLOPs. Finally, the self-critical sequence training (SCST) fine-tuning stage demonstrates an average iteration time of 1.629s across 5,265 samples, with reduced memory footprint of approximately 12GB, and exhibits a per-sample FLOPs of 13.92 GFLOPs and a per-iteration FLOPs of 27.84 GFLOPs, all evaluated on a single RTX4090 GPU.A more detailed comparison will be provided in the Appendix.
\begin{table}[htbp]
    \centering
    \footnotesize
    \setlength{\tabcolsep}{5pt} 
    \begin{tabular}{lccc}
    \toprule
    \textbf Configuration Item & \textbf{Pretrain} & \textbf{MLE} & \textbf{SCST} \\
    \midrule
    Dataset & ScanNet & ScanRefer & ScanRefer \\
    \hline
    Batch Size & 8 & 8 & 2 \\
    \hline
    Learning Rate & 5e-4 & 1e-4 & 1e-6 \\
    \hline
    Max Epochs & 1080 & 720 & 180 \\
    \hline
    GPU Memory Usage & \textasciitilde18GB & \textasciitilde18GB & \textasciitilde12GB \\
    \hline
    Captioner & None & Train & Train \\
    \hline
    Detector & Train & Train & Frozen \\
    \hline
    Beam Search & Not Used & Used &  Used \\
    \bottomrule
    \end{tabular}
    \caption{
        Configuration Comparison of Different Training Stages
    }
    \label{tab:training_configs}
\end{table}

\subsection{Comparison with Existing Methods}
We benchmark CAC($\mathcal{O}$) and CAC($\mathcal{O}\&\mathbb{H}$) against leading methods on ScanRefer~\cite{chen2020scanrefer} and Nr3D~\cite{achlioptas2020referit3d}. Evaluation applies standard IoU thresholds: 0.25/0.50 for ScanRefer (Table~\ref{tab:scanrefer}) and 0.50 exclusively for Nr3D (Table~\ref{tab:nr3d}), with C@0.5 as the primary benchmark. Baseline metrics reflect originally reported values, with ``\textit{-}" indicating unreported results.

Our multi-stage attention mechanism outperforms existing methods on both ScanRefer (requiring comprehensive attribute/spatial descriptions) and Nr3D (testing free-form human captions), achieving $\uparrow$2.7\% CIDEr@0.5 on ScanRefer and $\uparrow$4.6\% CIDEr@0.5 on Nr3D, the highest reported scores to date. This demonstrates significant improvements in diverse, accurate caption generation (Tables~\ref{tab:scanrefer},~\ref{tab:nr3d}) while validating real-world applicability.

\begin{figure*}
\centering
\includegraphics[width=1\linewidth]{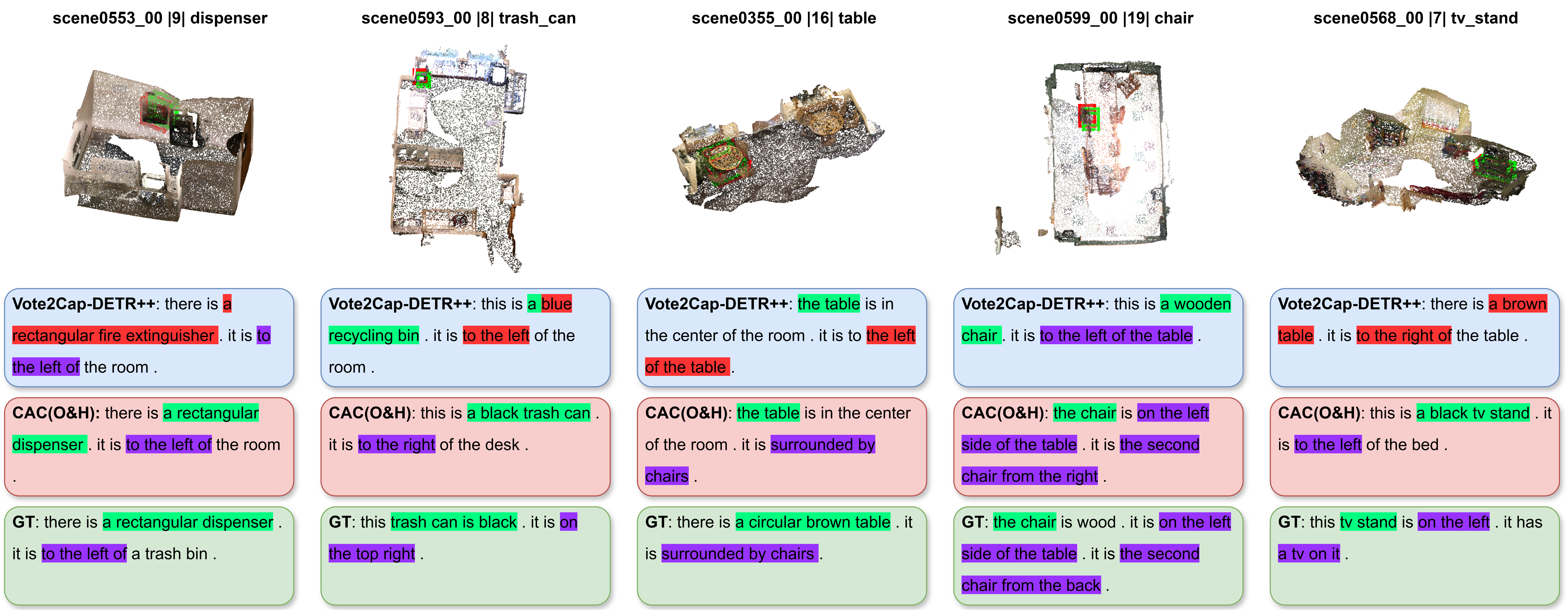}
\caption{
Qualitative results on the ScanRefer benchmark. Green highlights indicate accurate object descriptions, purple highlights denote contextual environmental descriptions, and red highlights identify erroneous descriptions with semantic deviations.
}
\label{fig:zhiliangfenxi} 
\end{figure*}

\input{tab/2_nr3d}
\input{tab/3_ablation}

\label{ablation}

\subsection{Ablation Study}

%
During our experimental investigation, we implemented BiCA's innovative approach of independently generating contextual and instance queries. However, the reproduced results proved unsatisfactory as detailed in Table~\ref{tab:cmpBiCA}, with comprehensive reproduction methodology documented in the Appendix. Consequently, we excluded BiCA from comparative analysis in Table~\ref{tab:scanrefer}.


Ablation studies on input \textbf{QKV} configurations within our bidirectional decoder framework demonstrate our approach's superiority over BiCA's methodology~\cite{kim2024bi}, evidenced by CAC(\ensuremath{\mathcal{O}})$^{BiCA}$ results in Table~\ref{tab:cmpBiCA}. Comprehensive ablation experiments on alternative input formulations during BiCA reproduction validate our architectural choices, with detailed analysis in the Appendix.
%
Comprehensive seed robustness testing on Nr3D across three random seeds (0, 333, 777) under MLE and SCST optimization confirms consistent performance (Table~\ref{tab:seed}). CAC($\mathcal{O}\&\mathbb{H}$) with SCST shows exceptional stability, with seed 777 achieving $51.18\% \pm 0.09\%$ (C@0.5), $29.58\% \pm 0.05\%$ (B-4@0.5), $26.28\% \pm 0.07\%$ (M@0.5), and $55.92\% \pm 0.06\%$ (R@0.5). The narrow confidence intervals demonstrate minimal variance and high reproducibility.

%% file: tab/2_nr3d.tex
\begin{table}[htbp]
    \centering

    \footnotesize 
    \setlength{\tabcolsep}{1mm} 
    \begin{tabular}{ccccccc}
    \toprule 
    \multirow{2}{*}{Method} & \multirow{2}{*}{$\mathcal{L}_{des}$}& \multicolumn{4}{c}{IoU = 0.50} \\ \cline{3-6}
    & & C@0.5$\uparrow$ & B-4@0.5$\uparrow$ & M@0.5$\uparrow$ & R@0.5$\uparrow$ \\ \hline
    Scan2Cap        & \multirow{11}{*}{MLE}  & 27.47           & 17.24             & 21.80           & 49.06           \\
    SpaCap3d        &                       & 33.71           & 19.92             & 22.61           & 50.50           \\
    D3Net           &                       & 33.85           & 20.70             & 23.13           & 53.38           \\
    Contextual&                       & 35.26           & 20.42             & 22.77           & 50.78           \\
    3DJCG           &                       & 38.06           & 22.82             & 23.77           & 52.99           \\
    V2C-DETR            &                   & 43.84  & 26.68    & 25.41  & 54.43  \\ 
    V2C-DETR++          &                       & 47.08   & 27.70   & 25.44  & 55.22 \\
    \textbf{V2C-DETR++$^{R}$}&
    & 46.10   & 27.19   & 25.38  & 55.22 \\
    \cline{1-1}
    \multicolumn{1}{l}{\textbf{\textit{Ours:}}} \\
    CAC(\ensuremath{\mathcal{O}})            &
    & \textbf{50.99}  & \textbf{28.89}  & \textbf{26.41}  & \textbf{56.18}  \\  
    CAC(\ensuremath{\mathcal{O}\&\mathbb{H}})          & 
    & 49.90  & 28.70  & 26.00  & 55.71  \\ 
    \hline
    %
    %
    %
    $\chi$-Tran2Cap & \multirow{9}{*}{SCST} & 33.62           & 19.29             & 22.27           & 50.00           \\
    Contextual&     & 37.37           & 20.96             & 22.89           & 51.11           \\
    D3Net         &                       & 38.42           & 22.22             & 24.74           & 54.37           \\
    V2C-DETR            &   
    & 45.53  & 26.88    & 25.43  & 54.76  \\ 
    V2C-DETR++          &                       & 47.62   & 28.41   & 25.63   & 54.77 \\
    \textbf{V2C-DETR++$^{R}$}&
    & 48.13   & 28.03   & 25.66  & 54.74 \\
    \cline{1-1}
    \multicolumn{1}{l}{\textbf{\textit{Ours:}}} \\
    CAC(\ensuremath{\mathcal{O}})            &
    & 50.99  & 29.32  & 26.13  & 55.81  \\ 
    CAC(\ensuremath{\mathcal{O}\&\mathbb{H}})          & 
    & \textbf{52.78}  & \textbf{29.78}  & \textbf{26.13}  & \textbf{55.94}  \\ 
    \bottomrule
    \end{tabular}
    \caption{
        Benchmark evaluation on the Nr3D validation set~\cite{achlioptas2020referit3d} demonstrates our method's superior performance against prior arts under both MLE training and SCST optimization frameworks.
    }
    \label{tab:nr3d}
\end{table}

\begin{table}[htbp]
    \centering
    
    \footnotesize 
    \setlength{\tabcolsep}{1mm} 
    \begin{tabular}{ccccccc}
    \toprule
    \multirow{2}{*}{Method} & \multirow{2}{*}{$\mathcal{L}_{des}$}& \multirow{2}{*}{Seed}& \multicolumn{4}{c}{IoU = 0.50} \\ \cline{4-7}
    &&& C@0.5$\uparrow$ & B-4@0.5$\uparrow$ & M@0.5$\uparrow$ & R@0.5$\uparrow$ \\ \hline
    \multirow{6}{*}{CAC(\ensuremath{\mathcal{O}})}                    & \multirow{3}{*}{MLE}
    &\textbf{0} & 50.99  & 28.89  & 26.41  & 56.18 \\  
            &    
    &\textbf{333}& 50.81   & 28.65   & 26.09   & 56.00   \\   
                                                                       & 
    &\textbf{777}& 50.64   & 28.50   & 26.29  & 56.05 \\   
    
    \cline{2-7}
                                                                    &  \multirow{3}{*}{SCST}
    &\textbf{0}& 50.99  & 29.32  & 26.13  & 55.81  \\  
            &    
    &\textbf{333}& 51.46   & 28.96   & 25.96   & 55.60   \\ 
                       & 
    &\textbf{777}& 51.12   & 28.91   & 26.01  & 55.50 \\  
    \hline
    \multirow{6}{*}{CAC(\ensuremath{\mathcal{O}\&\mathbb{H}})}                    & \multirow{3}{*}{MLE}
    &\textbf{0}& 49.90  & 28.70  & 26.00  & 55.71 \\  
            &    
    &\textbf{333}& 50.07   & 28.97   & 26.03   & 56.16   \\  
                       & 
    &\textbf{777}& 51.08   & 29.21   & 26.22  & 55.91 \\   
    \cline{2-7}
                                                                    &  \multirow{3}{*}{SCST}
    &\textbf{0}& 52.78  & 29.78  & 26.13  & 55.94  \\  
            &    
    &\textbf{333}& 49.34   & 28.98   & 26.16   & 56.12   \\  
                       & 
    &\textbf{777}& 53.18   & 29.58   & 26.28  & 55.92 \\    
    \bottomrule
    \end{tabular}
    \caption{
        Based on the Nr3D dataset~\cite{achlioptas2020referit3d}, stability test results across training seeds of 0, 333, and 777 are reported under both the MLE and SCST frameworks.
    }
    \label{tab:seed}
\end{table}

%% file: tab/3_ablation.tex
\begin{table}[htbp]
    \centering
    
    \footnotesize 
    \setlength{\tabcolsep}{1mm} 
    \begin{tabular}{cccccc}
    \toprule
    \multirow{2}{*}{Method} & \multirow{2}{*}{$\mathcal{L}_{des}$}& \multicolumn{4}{c}{IoU = 0.50} \\ \cline{3-6}
    && C@0.5$\uparrow$ & B-4@0.5$\uparrow$ & M@0.5$\uparrow$ & R@0.5$\uparrow$ \\ \hline
    \textbf{BiCA$^{R}$}                     & \multirow{5}{*}{MLE}
    & 65.22   & 37.59   & 26.87  & 55.76 \\
    \textbf{V2C-DETR++$^{R}$}          &    
    & 66.06   & 36.93   & 26.76   & 55.39   \\
    \textbf{CAC(\ensuremath{\mathcal{O}})$^{BiCA}$}                     & 
    & 67.07   & 36.27   & 26.49  & 54.98 \\
    CAC(\ensuremath{\mathcal{O}})            &
    & 68.07   & 36.53   & 26.72   & 55.08 \\  
    CAC(\ensuremath{\mathcal{O}\&\mathbb{H}})          & 
    & \textbf{69.92}   & \textbf{37.67}   & \textbf{26.89}   & \textbf{55.62}   \\
    \hline
    \textbf{BiCA$^{R}$}              & \multirow{5}{*}{SCST}
    & 76.42   & 38.75   & 26.70  & 54.66 \\
    \textbf{V2C-DETR++$^{R}$}          &                                      
    & 87.52   & 44.00   & 28.71   & 59.59  \\
    \textbf{CAC(\ensuremath{\mathcal{O}})$^{BiCA}$}                     & 
    & 77.86   & 39.10   & 27.01  & 54.83 \\
    CAC(\ensuremath{\mathcal{O}})            &
    & 79.09   & 38.96   & 26.85   & 54.95  \\ 
    CAC(\ensuremath{\mathcal{O}\&\mathbb{H}})          & 
    & \textbf{80.35}   & \textbf{39.95}   & \textbf{26.94}   & \textbf{55.66}  \\ 
    \bottomrule
    \end{tabular}
    \caption{
        Comparative evaluation against BiCA~\cite{kim2024bi} under identical experimental conditions: IoU threshold = 0.50.
    }
    \label{tab:cmpBiCA}
\end{table}

%% file: sec/6_quality.tex
\section{Qualitative Analysis}

%


Qualitative comparisons reveal CAC($\mathcal{O}\&\mathbb{H}$)'s significant descriptive improvements over Vote2Cap-DETR++ in accurately capturing object attributes and spatial relationships, as shown in Figure~\ref{fig:zhiliangfenxi}. Where Vote2Cap++ misidentifies a black trash can as a ``blue recycling bin" and mislabels a dispenser as a ``fire extinguisher", CAC correctly identifies these as a ``black trash can" and ``rectangular dispenser" respectively. More crucially, CAC demonstrates superior spatial awareness, describing furniture arrangements with precise references like ``it is the second chair from the right" and contextualizing relationships as ``the table is surrounded by chairs", while Vote2Cap++ generates implausible statements such as ``the table is to the left of the table" and vague descriptions like ``it is to the left of the room". 
As evidenced in Figure~\ref{fig:cidreComp}, our approach achieves earlier convergence to optimal solutions compared to Vote2Cap++, with dual-space learning in both Oblique and Lorentz Space yielding significantly faster convergence than Oblique-only optimization.
%


%% file: sec/7_conclu.tex
\section{Conclusion}


We introduce the \textbf{CAC} framework, which resolves the localization-contextualization conflict in 3D scene understanding by unifying Oblique-space geodesic attention—enforcing column-wise unit-norm constraints for isotropic stability—and Lorentzian hyperbolic attention ($\kappa_{\mathbb{H}} = -c$) for hierarchical semantics. This geometric synergy prevents dimensional collapse and ensures robust feature learning. State-of-the-art results on ScanRefer and Nr3D benchmarks demonstrate significant gains in localization accuracy and descriptive richness, laying a solid foundation for embodied AI applications.

\begin{figure}
\centering
\includegraphics[width=1\linewidth]{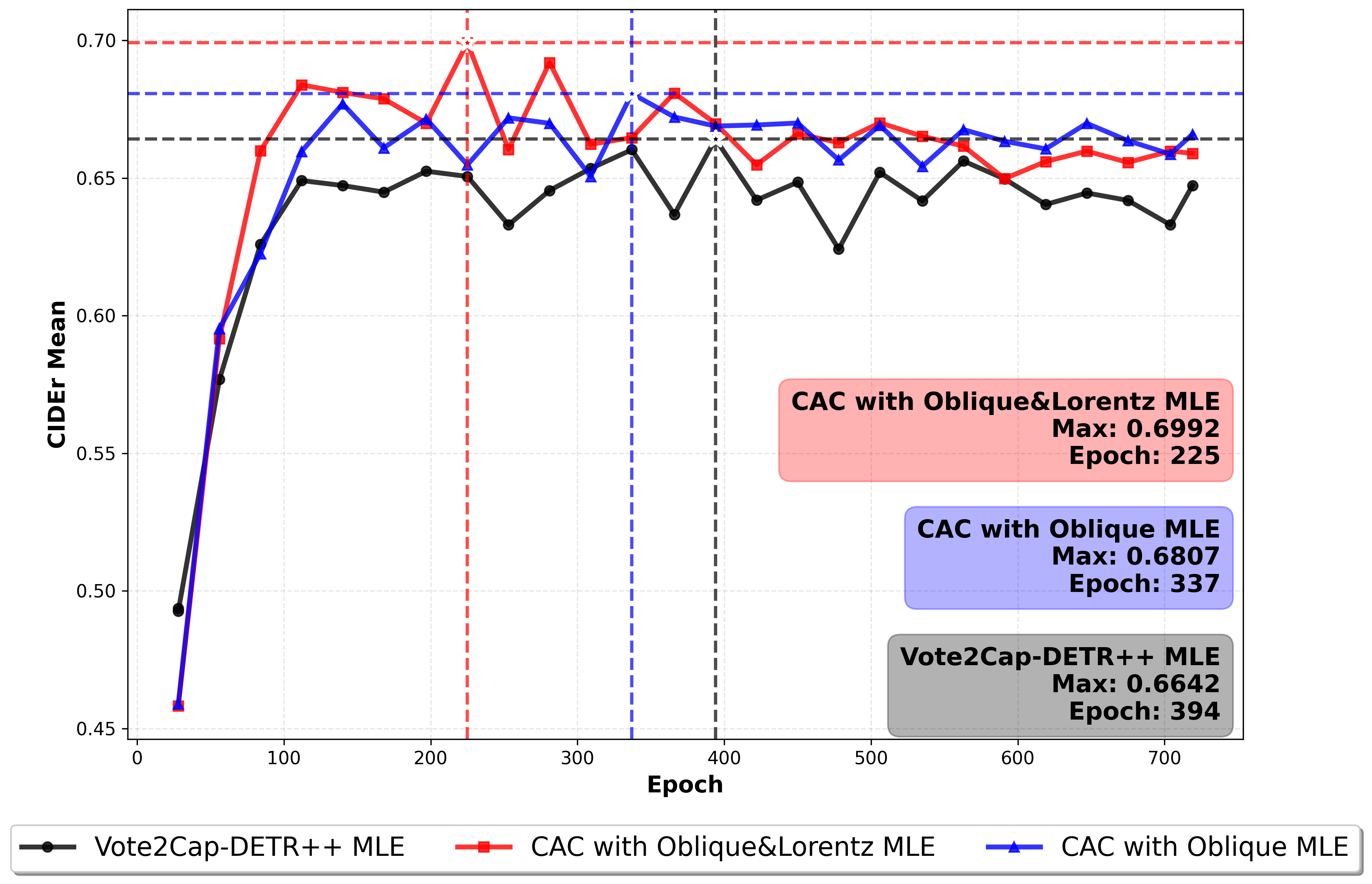}
\caption{
Comparison of CIDEr@0.5 Mean evaluation results of three experiments.}
\label{fig:cidreComp} 
\end{figure}

%% file: sec/8_Acknowledgments.tex
\section{Acknowledgments}

This research is supported by the General Program of Shanghai Natural Science Foundation (No.24ZR1419800, No.23ZR1419300), the National Natural Science Foundation of China (No.42130112), the Ministry of Industry and Information Technology of China, Science and Technology Commission of Shanghai Municipality (No.22DZ2229004), and Shanghai Frontiers Science Center of Molecule Intelligent Syntheses.